\documentclass[letterpaper, 10 pt, conference]{ieeeconf}

\IEEEoverridecommandlockouts
\usepackage{graphicx}

\usepackage{cite}
\usepackage{amsmath,amssymb,amsfonts}
\usepackage{algorithmic}
\usepackage{textcomp}
\usepackage{xcolor}
\usepackage{balance}
\usepackage{comment}
\usepackage[ruled,vlined]{algorithm2e}
\usepackage{array}
\usepackage{subcaption}
\usepackage{arydshln}
\usepackage{tensor}

\newcommand{\ie}{\emph{i.e.},}

\usepackage{float}
\usepackage{rotating}
\usepackage{nicematrix}  %
\usepackage{mathtools}
\usepackage{amsmath}
\usepackage{pifont}
\usepackage{cuted}
\usepackage{dblfloatfix}
\usepackage{gensymb}
\usepackage[ruled,vlined]{algorithm2e}
\usepackage{titlesec}
\titlespacing*{\section}{0pt}{.5em}{0pt}

\definecolor{LightCyan}{rgb}{0.88,1,0.88}
\definecolor{linear_color}{RGB}{220,223,240}
\definecolor{gray_bbox_color}{RGB}{243,243,244}

\definecolor{rebuttal}{rgb}{0,0,1}

\def\eqref#1{Eq.~(\ref{#1})}

\DeclareMathOperator*{\argmin}{argmin}

\newcommand{\methodname}{SOLVR}%
\newcommand{\crossmodalterm}{LiDAR-Visual}

\def\BibTeX{{\rm B\kern-.05em{\sc i\kern-.025em b}\kern-.08em
    T\kern-.1667em\lower.7ex\hbox{E}\kern-.125emX}}
\begin{document}

\title{\LARGE \bf \methodname{}: Submap Oriented LiDAR-Visual Re-Localisation
}
\author{Joshua Knights$^{1,2,3}$, Sebastián Barbas Laina$^1$, Peyman Moghadam$^{2,3}$, Stefan Leutenegger$^1$   
\thanks{$^{1}$ Smart Robotics Lab, School of Computation, Information and Technology, Technical University of Munich. Emails:  {\tt\footnotesize \emph{firstname.lastname@tum.de}}}
\thanks{$^{2}$ CSIRO Robotics, DATA61, CSIRO, Australia. E-mails: {\tt\footnotesize \emph{firstname.lastname}@csiro.au}} 
\thanks{$^{3}$ Queensland University of Technology (QUT), Brisbane, Australia.}
}

\bstctlcite{IEEEexample:BSTcontrol}

\maketitle
\begin{strip}
    \vspace{-3cm}
\end{strip}

\begin{abstract}
This paper proposes \methodname{}, a unified pipeline for learning based \crossmodalterm{} re-localisation which performs place recognition and 6-DoF registration across sensor modalities.  We propose a strategy to align the input sensor modalities by leveraging stereo image streams to produce metric depth predictions with pose information, followed by fusing multiple scene views from a local window using a probabilistic occupancy framework to expand the limited field-of-view of the camera.  Additionally, \methodname{} adopts a flexible definition of what constitutes positive examples for different training losses, allowing us to simultaneously optimise place recognition and registration performance. Furthermore, we replace RANSAC with a registration function that weights a simple least-squares fitting with the estimated inlier likelihood of sparse keypoint correspondences, improving performance in scenarios with a low inlier ratio between the query and retrieved place.
Our experiments on the KITTI and KITTI360 datasets show that \methodname{} achieves state-of-the-art performance for \crossmodalterm{} place recognition and registration, particularly improving registration accuracy over larger distances between the query and retrieved place.\looseness=-1

\end{abstract}

\section{Introduction}
\label{sec:intro}

Re-localisation is a fundamental challenge for mobile robotics, in which a robot localises itself into an existing map by coarsely identifying a similar place (\textit{place recognition}) before predicting the relative 6 Degree of Freedom (DoF) transform between the current view and the coarse retrieved place candidate (\textit{registration}). 
This task has been most commonly explored to date with vision \cite{galvez2012bags,hausler2021patch,sarlin2019coarse} or LiDAR \cite{uy2018pointnetvlad,vidanapathirana2022logg3d,komorowski2021egonn} sensing modalities, each of which present their own advantages and disadvantages.  Cameras are lightweight and inexpensive but struggle significantly with lighting and seasonal changes, as well as reverse revisits due to their limited field-of-view.  Conversely, LiDAR sensors are robust against lighting and seasonal changes while providing a panoptic view of the scene but are significantly heavier, more expensive and power-hungry than cameras.  

\crossmodalterm{} re-localisation aims to leverage the advantages of both sensor modalities by allowing for inter-modality place recognition and registration, enabling more flexibility for sensor setups when re-deploying into previously mapped environments or for exploration with swarms of heterogenous robots with different sensor capabilities and requirements.
However learning reliable cross-modal features which can be used to re-localise across sensor modalities is a non-trivial task, and the majority of previous approaches have either focused on the place recognition \cite{Cai2024VOLocVP,xu2024c2l,zhao2023attention,zheng2023i2p,shubodh2024lip,yin2021i3dloc,lee20232} or registration \cite{cattaneo2019cmrnet,li2021deepi2p,zhou2024differentiable, r3loc2023} aspects of the problem rather than construct a unified re-localisation pipeline. \looseness=-1

\begin{figure}[t]
    \centering
    \includegraphics[width=0.49\textwidth]{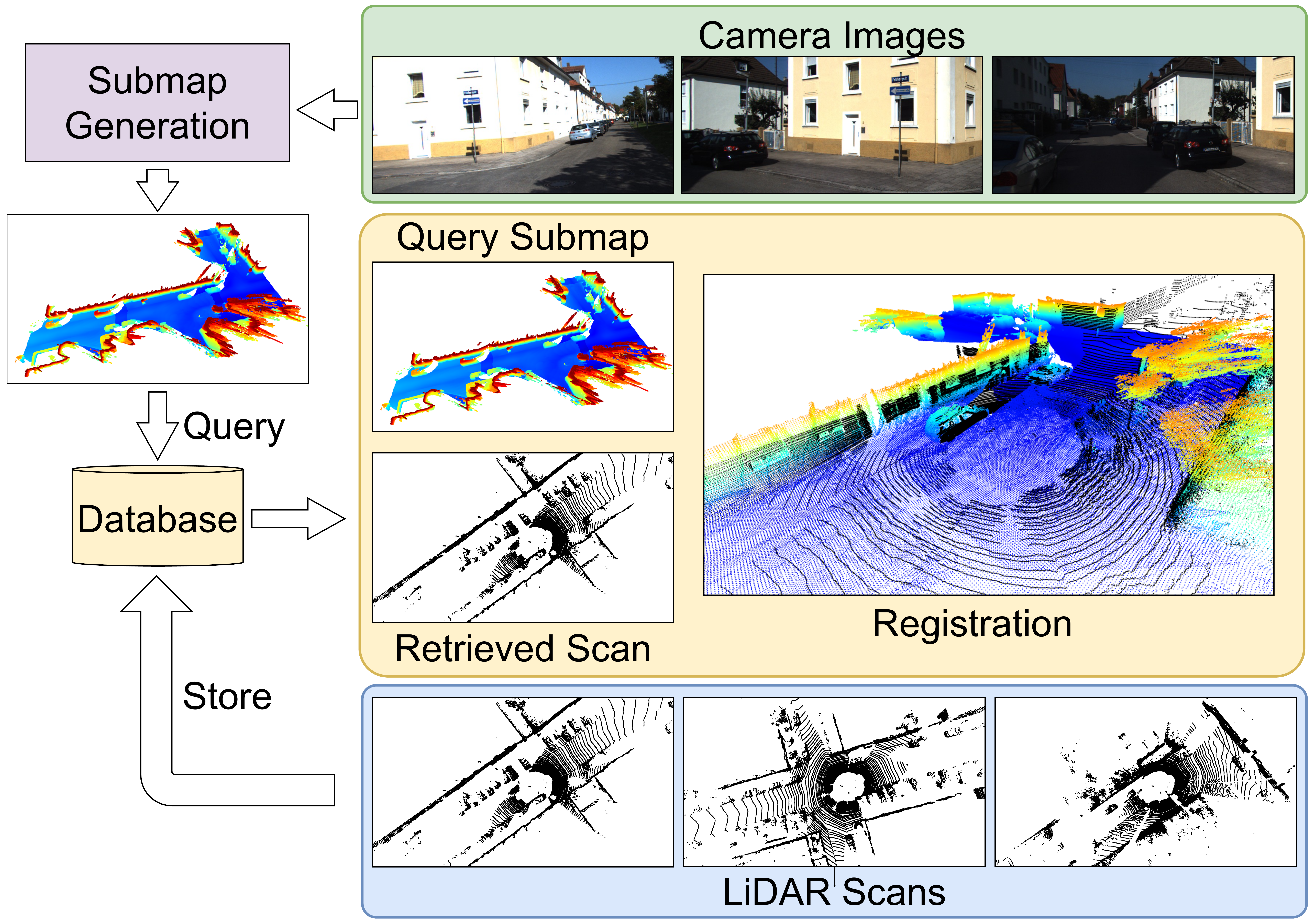}
    \caption{An illustration of our approach to the task of \crossmodalterm{} re-localisation.  \methodname{} constructs 3D submaps from a stream of input camera images in order to align the camera and LiDAR sensor modalities.  The submap is used to retrieve a corresponding place from a database of LiDAR scans, at which point the submap and scan are registered in order to estimate the current sensor pose.}
    \label{fig:overview}
    \vspace{-5mm}
\end{figure}

In this paper we propose \methodname{}, a unified pipeline for learning-based \crossmodalterm{} re-localisation.  \methodname{} constructs 3D submaps from local windows of incoming camera frames by predicting and probabilistically fusing the metric depth from pairs of stereo images, aligning the sensor modalities and expanding the effective field of view of the camera sensor for more reliable re-localisation.  An overview of the system operation is shown in Figure \ref{fig:overview}.  We also propose a novel approach for defining positive examples across the different aspects of re-localisation to simultaneously optimise place recognition and registration performance in our pipeline during training, and employ a lightweight adaption of the keypoint registration approach introduced by \cite{bai2021pointdsc} for dramatically increased speed and reliability of the sensor pose estimation for difficult registration examples.  In short, the primary contributions of this paper are as follows:\looseness=-1

\begin{itemize}
    \item We propose \methodname{}, a unified pipeline for learning-based \crossmodalterm{} re-localisation.
    \item We propose a novel pipeline for generating 3D submaps from the camera image stream by predicting the metric depth for each incoming stereo image pair and probabilistically integrating frames from a local window for cleaner, geometrically consistent submaps.
    \item We propose a novel approach for defining positive examples across the different aspects of re-localisation to simultaneously optimise place recognition and registration performance in our pipeline.
    \item We employ a lightweight registration approach to achieve faster and more robust pose estimation than the RANSAC baseline. \looseness=-1
    \item We benchmark our approach on the KITTI \cite{geiger2013kittidataset} and KITTI360 \cite{liao2022kitti360} datasets, and demonstrate consistent state-of-the-art results compared to existing \crossmodalterm{} place recognition and registration approaches.  In particular, we show that \methodname{} dramatically improves registration accuracy for larger distances between the query and retrieved place in the real world.
\end{itemize}

\section{Related Work}
\label{sec:rel_work}

\subsection{\crossmodalterm{} Place Recognition}
\label{sec:rel_work_pr}

With some exceptions \cite{zhao2023attention}, the existing literature on \crossmodalterm{} place recognition can be broadly categorised into two families of approaches: 2D alignment, where the LiDAR scan is converted to a 2D range \cite{shubodh2024lip,yin2021i3dloc,lee20232} or bird's eye view \cite{zheng2023i2p} image for matching, and 3D alignment \cite{xu2024c2l,zheng2023i2p,Cai2024VOLocVP}, where a 3D construction of the scene is derived from the camera images in order to align the sensor modalities.  While approaches following the 2D alignment strategy have achieved some limited success, they face two key drawbacks.  Firstly the panoptic field-of-view of a LiDAR sensor is much broader than a typical perspective camera, with the common solution to this issue \cite{xu2024c2l,zhao2023attention,shubodh2024lip,lee20232} being to crop the field-of-view of the 2D LiDAR image to match that of the camera, sacrificing a key advantage of the LiDAR sensor modality and degrading performance for reverse and orthogonal revisits.  Secondly, projecting 3D points into a 2D image flattens the rich geometric information provided by the LiDAR point cloud, once again sacrificing a key advantage of the sensor modality.
More recently, several approaches have proposed aligning the sensor modalities by constructing 3D views of a scene from either reconstruction with SLAM \cite{Cai2024VOLocVP} or depth prediction \cite{xu2024c2l,zheng2023i2p}.  While these approaches do a better job of allowing for the 3D geometry of the LiDAR point cloud to be leveraged, 3D reconstructions from a single frame still suffer from the field-of-view disparity between the LiDAR and a perspective camera.  In order to address both of these challenges \methodname{} fuses multiple dense depth predictions in a local window to create large 3D submaps from stereo camera frames, allowing for \crossmodalterm{} place recognition while leveraging both the panoptic view or rich 3D geometry provided by the LiDAR sensor modality. \looseness=-1

\subsection{\crossmodalterm{} Registration}

Compared to place recognition the existing literature for \crossmodalterm{} registration places less of a focus on modality alignment, with most approaches predicting the 6-DoF transformation between the raw 2D image and 3D LiDAR scan.  CMRNet \cite{cattaneo2019cmrnet} uses a modified optical flow prediction network in combination with a coarse-to-fine iterative refinement scheme to predict the camera pose as network output; however, it requires an initial rough alignment proposal between the image and corresponding LiDAR scan.  DeepI2P \cite{li2021deepi2p} performs registration in a correspondence-free fashion by classifying the overlapping region between the image and point cloud and using the inverse camera projection to predict the camera pose.  Finally, CorrI2P \cite{ren2022corri2p} and VP2P-Match \cite{zhou2024differentiable} predict dense correspondences between pixels and points using intersection detection to remove outliers before solving for the pose with a combination of EPnP \cite{lepetit2009ep} and RANSAC \cite{fischler1981random}; however, as we show in Section \ref{sec:reg_performance} the performance of these approaches rapidly degrades as the distance between the camera and LiDAR sensors increases.

\subsection{Unified Re-Localisation}
Methods performing unified re-localisation for image \cite{sattler2016efficient,Sarlin2018FromCT} and LiDAR \cite{komorowski2021egonn,du2020dh3d,zhang2023instaloc, spectralgv} adopt a two-stage approach, where for a given query a place recognition module first coarsely retrieves a similar place from a database before predicting the metric pose of the sensor in the environment by predicting the 6-DoF registration of the query to the coarse retrieved place.  R3Loc \cite{r3loc2023} leverages multi-modal information for hypothesis verification when re-localising, but still requires both sensor modalities to be present on the re-localising platform.  Concurrent to this work, \cite{zhang2024visual} propose a approach for visual re-localisation in 3D maps by synthesising image frames from point cloud, mesh or NeRF representations of the environment to be used as a database for visual queries.  Instead of synthesising limited field-of-view 2D images from a 3D map, \methodname{} instead generates large 3D submaps from camera frames in order to leverage the full advantages of both sensor modalities.

\section{Methodology}

\subsection{Problem Description}
 Assume we have a set of greyscale stereo images $\mathcal{I} = \left\{\left(\mathcal{I}_{1}^{\text{left}}, \mathcal{I}_{1}^{\text{right}}\right),\left(\mathcal{I}_{2}^{\text{left}}, \mathcal{I}_{2}^{\text{right}}\right), \ldots,\left(\mathcal{I}_{T}^{\text{left}}, \mathcal{I}_{T}^{\text{right}}\right) \right\}$, where $\mathcal{I}_{i}^{\text{left}}, \mathcal{I}_{i}^{\text{right}} \in \mathbb{R}^{H\times W}$ denote the left and right image of stereo pair $i$ with height $H$ and width $W$ and $T=\left|\mathcal{I}\right|$,  with corresponding 6-DoF camera trajectory $\mathcal{T} = \left\{\mathcal{T}_1, \mathcal{T}_2, \ldots, \mathcal{T}_T\right\}$ where $\mathcal{T}_i = \left(\mathbf{R}_i, \mathbf{t}_i\right)$ denotes the camera pose for stereo pair $i$ consisting of rotation $\mathbf{R}_i$ and translation $\mathbf{t}_i$. The goal of \crossmodalterm{} re-localisation is to, for a given \textit{query} from the image set $\mathcal{I}$, retrieve a corresponding \textit{candidate} from a pre-existing set of $\hat{T}$ 3D LiDAR scans $\mathcal{P} = \left\{\mathcal{P}_1, \mathcal{P}_2, \ldots, \mathcal{P}_{\hat{T}}\right\}$ from the same environment, where $\mathcal{P}_i \in \mathbb{R}^{N\times 3}$ denotes point cloud $i$ with $N$ points, and predict the 6-DoF registration of the  query to the candidate.

\subsection{Overview}

In this section, we provide an overview of \methodname{}.  Firstly, we use stereo depth estimation to create a 3D depth projection for each query image, before fusing them into larger refined submaps using the camera trajectory and occupancy mapping.  Secondly, we train two sparse convolutional networks~-- one for each input modality~-- to extract global feature embeddings, keypoint coordinates, and keypoint features for each input submap or scan.  Finally, we use the global features to perform place recognition by retrieving the closest candidate for a given query from a database of previously observed places and use the local keypoint co-ordinates and features to predict the 6-DoF transform between the query and candidate using our proposed registration.

\subsection{Submap Generation}
\label{sec:submap_generation}

\begin{figure}[t]
    \centering
    \includegraphics[width=\linewidth]{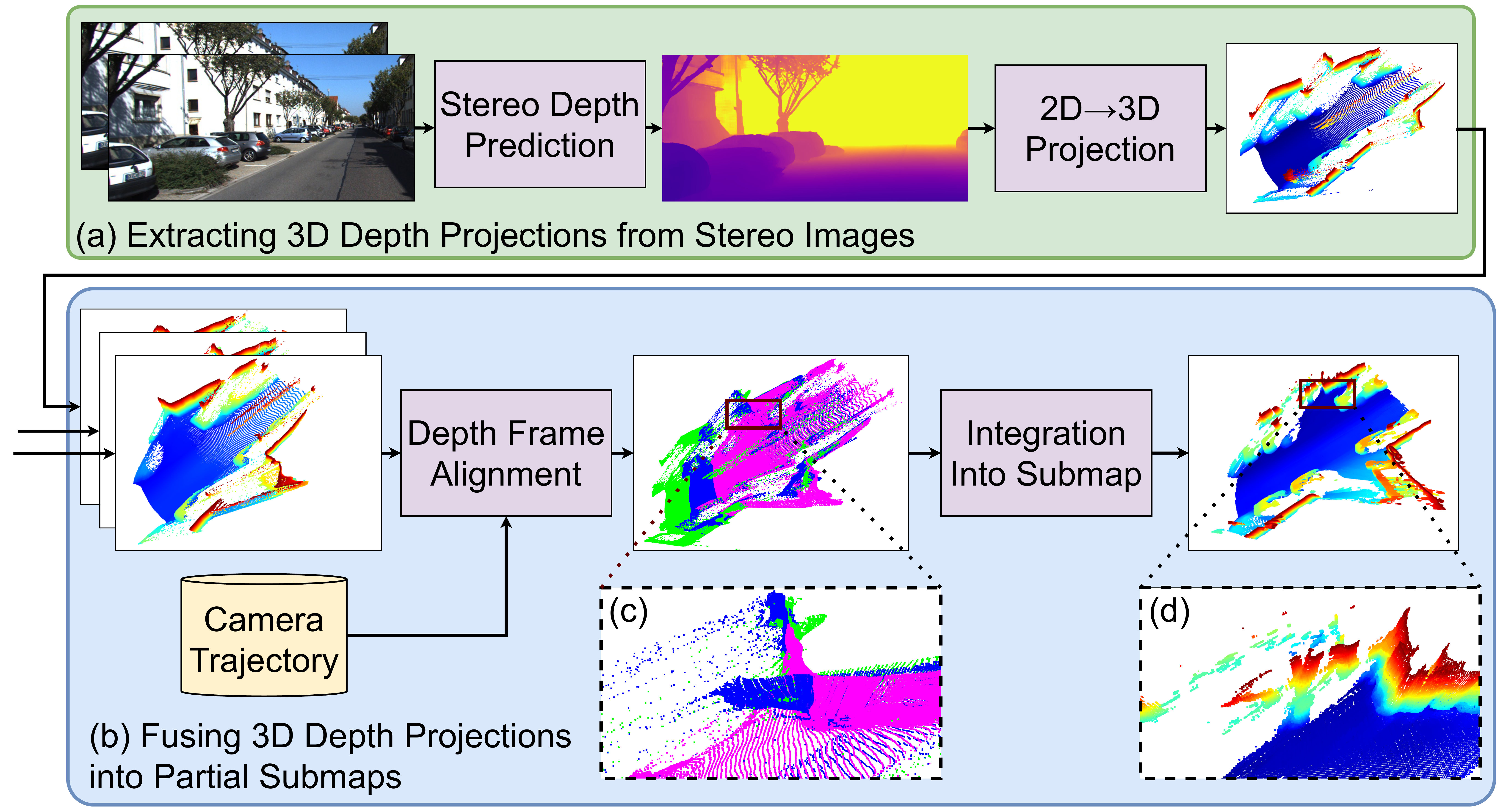}
    \caption{Partial submap generation pipeline.  For each pair of stereo images, we create a 3D depth projection by predicting the metric depth using a neural network and projecting each pixel into 3D using the camera intrinsics.  We then align the depth projections from a local window using the camera trajectory and fuse them using a Bayesian occupancy update to construct our partial submap.  We note that the noise in zoomed-in region (c) is significantly reduced in (d), due to the Bayesian occupancy update performed when integrating individual depth projections into the partial submap.}
    \label{fig:submap_generation}
    \vspace{-5mm}
\end{figure}

As discussed in Section \ref{sec:rel_work_pr}, two of the main challenges posed by the task of \crossmodalterm{} re-localisation are the significant gap between the input modalities and the limited field-of-view of the query images versus the LiDAR scan.  In order to jointly address these challenges, we propose an approach for constructing dense 3D submaps from sequences of input images to improve the similarity of the query and candidate inputs both in terms of modality and proportion of the real-world scene represented in the input.

For each stereo image pair $\mathcal{I}_{i}^{\text{left}},  \mathcal{I}_{i}^{\text{right}}$ we use a depth prediction network to predict the metric depth of the input before projecting each pixel to a 3D point using the camera intrinsics to form depth projection $\mathcal{D}_i \in \mathbb{R}^{\left(H\times W\right)\times 3}$.  We use Unimatch \cite{xu2023unifying} for depth prediction, using pre-trained checkpoints supplied by the authors.  We then align local windows of depth projections using the camera trajectory before fusing them to create a set of partial submaps $\hat{\mathcal{S}} = \left\{\hat{\mathcal{S}}_1, \hat{\mathcal{S}}_2, \ldots, \hat{\mathcal{S}}_K \right\}$, where $K$ denotes the number of partial submaps produced.  We integrate incoming depth projections into the partial submap $\hat{\mathcal{S}}_i$ as long as the volumetric intersection of the incoming depth projection with partial submap $\hat{\mathcal{S}}_{i-1}$ is greater than $20\%$, or as long as there are less than 10 depth projections integrated into $\hat{\mathcal{S}}_i$.  Figure \ref{fig:submap_generation} provides a visual overview of the partial submap generation process.  Finally, we use the camera poses for each partial submap to combine a sliding window of $7$ partial submaps into a set of complete 3D submaps $\mathcal{S}$, which we use for training and evaluation. The poses will in practice be obtained from a visual or multi-sensor odometry or SLAM system.

We observe that naively fusing the depth projections results in noticeably noisy submaps, due to the aggregation of artifacts and minor geometric inconsistencies between the subsequent depth projections.  Therefore, we use the occupancy mapping supereight2 \cite{funk2021multi} to sequentially integrate each new depth projection into its corresponding partial submap. Supereight2 performs a Bayesian update of the occupancies from different depth estimates, alleviating the geometric inconsistencies that emerge in the naive aggregation. Our experiments demonstrate that this approach significantly reduces the noisiness of our generated submaps, and leads to substantially improved \crossmodalterm{} re-localisation.

\subsection{Network Architecture and Training}
\methodname{} employs two parallel network backbones, for extracting features from the query submaps $\mathcal{S}$ and candidate LiDAR scans $\mathcal{P}$ respectively.  Following EgoNN \cite{komorowski2021egonn}, our network design consists of a sparse convolutional encoder followed by a global encoder which outputs a global embedding $\mathbf{g} \in \mathbb{R}^{256}$ and a local encoder which outputs keypoint features, coordinates, and saliency, denoted by $\mathbf{l}\in\mathbb{R}^{N\times128}, \mathbf{q}\in\mathbb{R}^{N\times 3}, \mathbf{\sigma}\in\mathbb{R}^{N}$ respectively, where here $N$ denotes the number of sparse keypoints.  For place recognition, we train the network using the triplet loss:
\vspace{-.5mm}
\begin{equation}\mathcal{L}_\text{Triplet}=\left[\left\|\mathbf{g}^a-\mathbf{g}^p\right\|-\left\|\mathbf{g}^a-\mathbf{g}^n\right\|+m\right]_+,
\end{equation}

where $\mathbf{g}^a, \mathbf{g}^p, \mathbf{g}^n$ denote the global embeddings produced by anchor submap $\mathcal{S}_a$ and corresponding positive and negative LiDAR scans $\mathcal{P}_p$ and $\mathcal{P}_n$ and $\left[\cdot\right]_+$ denotes the function $\max\left(\cdot, 0\right)$.  Next, following \cite{komorowski2021egonn} we combine three loss terms to train the local encoder.  Firstly, for locally consistent keypoint features we use a feature descriptor loss:
\begin{equation}
    \mathcal{L}_{\text{Desc}} = -\frac{1}{N'}\sum^{N'}\ln\left(\frac{\exp\left(
        \left\|\mathbf{l}^a_i- \mathbf{l}^p_{\text{nn}\left(i\right)}\right\|\right)}{\sum^M_{j=1}\exp\left(
        \left\|\mathbf{l}^a_i- \mathbf{l}^p_j\right\|\right)}\right),
\end{equation}

where $\mathbf{l}^a_i$ denotes the $i$-th local feature from submap $\mathcal{S}_a$ with a corresponding keypoint in scan $\mathcal{P}_p$, $\mathbf{l}^p_{\text{nn}\left(i\right)}$ denotes a corresponding feature to $\mathbf{l}^a_i$ in $\mathcal{P}_p$, and $N'$ and $M$ denote the number of keypoints in $\mathcal{S}_a$ with correspondences in $\mathcal{P}_p$ and the number of keypoints in $\mathcal{P}_p$ respectively.  To train keypoint co-ordinates, we use the probabilistic chamfer loss \cite{li2019usip}:
\begin{equation}
    \mathcal{L}_{\text{PC}} = \sum^N_{i=1}\left(\ln s_i^{a,p} + \frac{d_i^{a,p}}{s_i^{a,p}} \right) + \sum^M_{j=1}\left(\ln s_j^{p,a} + \frac{d_j^{p,a}}{s_j^{p,a}} \right),
\end{equation}

where $d_i^{a,p}$ is shorthand for $\left\|\mathbf{q}^a_i-\mathbf{q}^p_{\text{nn}\left(i\right)}\right\|$ and $s_i^{a,b}$ is shorthand for $\frac{1}{2}\left(\sigma^a_i+\sigma^p_{\text{nn}\left(i\right)}\right)$, and point-to-point loss:
\begin{equation}
    \mathcal{L}_{\text{P2P}} = \sum_{i=1}^N \min_{\mathbf{p}^a_j\in \mathcal{P}^a}\left\|\mathbf{q}^a_i-\mathbf{p}^a_j\right\| + \sum_{j=1}^M \min_{\mathbf{p}^b_i\in \mathcal{P}^b}\left\|\mathbf{q}^b_j-\mathbf{p}^b_i\right\|,
\end{equation}

where $\mathbf{p}_i^a$ denotes the $i$-th point in the input submap or scan $\mathcal{S}_a$ or $\mathcal{P}_a$.  These three losses train the local keypoint encoder to produce sparse locally consistent keypoint coordinates and features that can be used for query and candidate registration.  Our total loss can be expressed as: 
\begin{equation}
    \mathcal{L}_{\text{Total}} = \mathcal{L}_{\text{Triplet}} + \mathcal{L}_{\text{Desc}} + \mathcal{L}_{\text{PC}} + \mathcal{L}_{\text{P2P}}.
\end{equation}

\subsection{Constructing Training Tuples}
The typical approach~\cite{hausler2021patch,komorowski2021minkloc3d,komorowski2022improving,vidanapathirana2022logg3d,komorowski2021egonn,du2020dh3d} to constructing training pairs for training a network on the task of place recognition or registration is to define distance thresholds $\alpha_{\text{pos}}$ and $\alpha_{\text{neg}}$ such that pairs of queries and candidates within $\alpha_{\text{pos}}$ distance of one another are considered positives and pairs outside of $\alpha_{\text{neg}}$ distance of one another are considered negatives.  In our proposed approach, we diverge from this convention in two ways.  Firstly, we use the inlier ratio of a given query and candidate pair as the metric by which we separate our positive and negative training examples, following the observations of GeoAdapt~ \cite{knights2023geoadapt} that this approach leads to stronger downstream place recognition performance.  
Secondly, we observe in our experiments that the place recognition component of our approach performs best when trained with similar (\ie{} high overlap) scans and submaps, as it learns to more discriminately identify similar inputs; conversely, the registration module learns to register difficult (\ie{} low overlap) re-visits much more reliably when trained with dificult positive examples which force the network to learn more robust local features.  This leads to a trade-off during training, where the optimal value of $\alpha_{\text{pos}}$ for either task leads to significantly reduced performance on the other.  
Therefore, instead we propose to use two separate positive thresholds $\alpha_{\text{pos}}^{\text{pr}}$ and $\alpha_{\text{pos}}^{\text{reg}}$ for constructing positive query-candidate pairs for the losses related to place recognition and registration respectively, which as we demonstrate in our experiments allows us to bypass this trade-off and simultaneously optimise performance in both tasks.

\subsection{Registration}
\label{sec:method_reg}
While RANSAC \cite{fischler1981random} is a powerful tool for outlier rejection and registration, it performs unreliably when registering query-candidate pairs with a low inlier ratio.  Therefore, in order to achieve more robust regsitration under challenging circumstances we employ a lightweight implementation of the registration approach proposed by \cite{bai2021pointdsc} where geometric consistency is leveraged to estimate the inlier probability of the proposed correspondences in order to filter outliers and weight inliers when solving for the relative transform.

Given local features $\mathbf{l}_a, \mathbf{l}_p$ and keypoints $\mathbf{q}_a, \mathbf{q}_p$ for query submap $\mathcal{S}_a$ and positive candidate scan $\mathcal{P}_p$ we construct a set of point correspondences $C^{a\leftrightarrow p}$, where each correspondence $i$ is denoted as $c_i \in C^{a\leftrightarrow p} = \left\{\left(\mathbf{q}^a_i,\mathbf{l}^a_i\right), \left(\mathbf{q}^p_i,\mathbf{l}^p_i\right )\right\}$ with $\mathbf{q}^a_i ,\mathbf{l}^a_i , \mathbf{q}^p_i ,\mathbf{l}^p_i$ denoting the keypoint and local feature from $\mathcal{S}_a$ and $\mathcal{P}_p$ linked by correspondence $c_i$.  From this we can construct a geometric consistency matrix $\mathbf{M}_{a,p} \in \mathbb{R}^{\left|\mathcal{S}_a\right|\times \left|\mathcal{P}_p\right|}$ where every member $m_{i,j}$ measure the pairwise length consistency between correspondences $c_i, c_j$ defined as:
\begin{equation}
    m_{i,j} = \left[1-\frac{d_{i,j}^2}{d_{\text{thr}}^2}\right]_+, ~d_{i,j} = \left|\left\|\mathbf{q}^a_i-\mathbf{q}^a_j\right\|_2 - \left\|\mathbf{q}^p_i-\mathbf{q}^p_j\right\|_2\right| , 
\end{equation}

where $d_{\text{thr}}$ is a hyperparameter which controls sensitivity to length difference.  Following \cite{spectralmatch}, we can then consider the leading eigenvector $\mathbf{e} \in \mathbb{R}^{\left| C^{a\leftrightarrow b} \right|}$ as the inlier probability of each correspondence $c_i$.  We can then use $\mathbf{e}$ as a weight to estimate the transformation through least-squares fitting:
\begin{equation}
    \mathbf{R}, \mathbf{t} = \argmin_{\mathbf{R},\mathbf{t}}\sum_{i | \mathbf{e}_i > \tau} ^{\left|C^{a\leftrightarrow b}\right|}\mathbf{e}_i\left\|\mathbf{R} \mathbf{q}^a_i + \mathbf{t} - \mathbf{q}^p_i\right\|,
\end{equation}

where $\mathbf{R}, \mathbf{t}$ are the rotation and translation of the predicted transform and $\tau$ is a hyperparameter used to filter out low-confidence correspondences from consideration.

\begin{table}[t]
    \centering
    \begin{NiceTabular}{l|c|c|c}
        \hline 
        \Block{2-1}{Sequence} & \Block{2-1}{Queries} & \Block{2-1}{Candidates} & Average Positives \\
        &&& Per Query \\
        \hline 
        KITTI-00 & 562 & 2900 & 58.65\\
        KITTI360-00 & 876 & 7513 & 51.29\\
        KITTI360-09 & 2537 & 5264 & 70.41 \\
        \hdashline 
        KITTI-09 & 1450 & 1591 & 37.44 \\
        KITTI-10 & 934 & 1201 & 54.48 \\
        \hline 
    \end{NiceTabular}
    \caption{Evaluation sequence statistics.  The dashed line separates sequences used to evaluate both place recognition and registration (above) and those only used to evaluate registration (below).}
    \label{tab:eval_info}
    \vspace{-5mm}
\end{table}

\section{Experiments}

\begin{table*}[t]
    \centering
    \begin{NiceTabularX}{\textwidth}{X|| cc|cc|cc|cc|cc|cc}
        \CodeBefore
            \rowcolor{green!10}{8}
        \Body 
        \hline 
        \Block{3-1}{Method}  &  \Block{1-4}{KITTI-00*} & & & & \Block{1-4}{KITTI360-00}  & & & & \Block{1-4}{KITTI360-09}  & & & \\
        & \Block{1-2}{5m} & & \Block{1-2}{20m} & & \ \Block{1-2}{5m} & & \Block{1-2}{20m} & & \Block{1-2}{5m} & & \Block{1-2}{20m} & \\
        & R@1$\uparrow$ & R@5$\uparrow$ & R@1$\uparrow$ & R@5$\uparrow$ & R@1$\uparrow$ & R@5$\uparrow$ & R@1$\uparrow$ & R@5$\uparrow$ & R@1$\uparrow$ & R@5$\uparrow$ & R@1$\uparrow$ & R@5$\uparrow$ \\
        \hline 
        ScanContext$^{\dag}$ \cite{kim2018scan}   & 7.6 & 8.0 & 7.7 & 9.4 & 7.5 & 7.9 & 9.4 & 10.5 & 9.0 & 9.8 & 10.0 & 11.5 \\
        EgoNN$^{\dag}$ \cite{komorowski2021egonn} & 49.1 & 60.0 & 54.8 & 68.7 & \underline{40.2} & \underline{64.4} & \underline{73.2} & \underline{84.2} & \underline{47.9} & \underline{64.3} & \underline{58.7} & \underline{70.7} \\
        LIPLoc \cite{shubodh2024lip} & 46.6 & 74.3 & 52.1 & 76.5 & 1.94 & 5.8 & 11.0 & 21.0 & 21.7 & 33.4 & 27.0 & 41.6 \\ %
        
        C2L-PR \cite{xu2024c2l} & \underline{71.3} & \underline{84.2} & \underline{78.0} & \underline{88.4} &--&--&--&--&--&--&--&-- \\
        \hdashline
       \methodname{} (Ours)   & \textbf{87.9} & \textbf{94.0} & \textbf{96.4} & \textbf{96.8} & \textbf{58.1} & \textbf{74.8} & \textbf{76.3} & \textbf{84.6} & \textbf{70.2} & \textbf{85.5} & \textbf{82.2} & \textbf{90.3}\\ %
        \hline 
    \end{NiceTabularX}
    \caption{\crossmodalterm{} Place Recognition results on the KITTI and KITTI360 datasets.  $^{\dag}$   Indicates methods originally proposed for unimodal LiDAR tasks that have been adapted to \crossmodalterm{} by using our generated submaps as queries.  * Indicates sequences where the OKVIS2 \cite{leutenegger2022okvis2} odometry was used to calculate the camera trajectory for submap construction.}
    \label{tab:pr_results}
\end{table*}

\begin{table*}[t]
    \centering
    \begin{NiceTabularX}{\textwidth}{X|| ccc|ccc|ccc}
        \CodeBefore
            \rowcolor{green!10}{5}
        \Body 
        \hline 
        \Block{2-1}{Method}  & \Block{1-3}{KITTI-00*} & & & \Block{1-3}{KITTI360-00}  & & & \Block{1-3}{KITTI360-09}  & & \\
        & Acc.\% $\uparrow$ & RRE (\degree) $\downarrow$ & RTE (m) $\downarrow$  & Acc.\% $\uparrow$ & RRE (\degree) $\downarrow$ & RTE (m) $\downarrow$   & Acc.\% $\uparrow$ & RRE (\degree) $\downarrow$ & RTE (m) $\downarrow$   \\
        \hline 
        EgoNN$^{\dag}$ \cite{komorowski2021egonn} &  97.1 & \underline{1.14} & \underline{0.58}  & 82.8 & 1.86 & 0.38 & 95.5 & \underline{1.29} & \underline{0.35}\\ %
        \hdashline 
        \methodname{}-RANSAC & \underline{99.1} & 1.30 & 0.63 & \underline{94.2} & \underline{1.83} & \underline{0.37} & \underline{99.1} & 1.35 & \underline{0.35}\\
        \methodname{} (Ours) & \textbf{99.4} & \textbf{0.73} & \textbf{0.53} & \textbf{100.0} & \textbf{1.26} & \textbf{0.28} & \textbf{99.7} & \textbf{0.82} & \textbf{0.28} \\
        \hline 
    \end{NiceTabularX}
    \caption{6-DoF \crossmodalterm{} Registration results for the \textit{Top-1} evaluation setup.  Registration accuracy is only reported for successful top-1 place retrievals, and the RRE and RTE reported is the mean of only successful registrations. }
    \label{tab:reg_top1_results}
\end{table*}

\begin{table*}[t]
    \centering
    \begin{NiceTabularX}{\textwidth}{X||ccc|ccc|ccc}
        \CodeBefore
            \rowcolor{green!10}{5}
        \Body 
        \hline 
        \Block{2-1}{Method}  & \Block{1-3}{KITTI-00*} & & & \Block{1-3}{KITTI360-00}  & & & \Block{1-3}{KITTI360-09} & & \\
        & Acc.\% $\uparrow$ & RRE (\degree) $\downarrow$ & RTE (m) $\downarrow$  & Acc.\% $\uparrow$ & RRE (\degree) $\downarrow$ & RTE (m) $\downarrow$   & Acc.\% $\uparrow$ & RRE (\degree) $\downarrow$ & RTE (m) $\downarrow$   \\
        \hline 
        EgoNN$^{\dag}$ \cite{komorowski2021egonn} & 66.6 & 24.41 & 5.53 & 44.9 & 45.29 & 8.21 & 54.0 & 36.17 & 7.07\\
        \hdashline 
        \methodname{}-RANSAC & \underline{80.9} & \underline{17.00} & \underline{3.61} & \underline{64.1} & \underline{36.57} & \underline{5.86} & \underline{72.4} & \underline{25.58} & \underline{4.56} \\
        \methodname{} (Ours) & \textbf{96.0} & \textbf{3.39} & \textbf{1.02} & \textbf{93.8} & \textbf{6.45} & \textbf{0.86} & \textbf{93.7} & \textbf{4.69} & \textbf{0.92} \\
        \hline 
    \end{NiceTabularX}
    \caption{6-DoF \crossmodalterm{} Registration results for the \textit{Comprehensive} evaluation setup.  Registration accuracy is reported over all possible positive pairs, and RRE and RTE reported is the mean of both successful and failed registrations.}
    \label{tab:reg_comp_results}
    \vspace{-5mm}
\end{table*}

\subsection{Datasets}
We benchmark \methodname{} on the KITTI~\cite{geiger2013kittidataset} and KITTI360~\cite{liao2022kitti360} datasets.  For all results except those reported in Table \ref{tab:reg_nopr} and Figure \ref{fig:success_vs_distance}, we train with KITTI sequences 01-10 and report results on sequences KITTI-00, KITTI360-00 and KITTI360-09.  For Table \ref{tab:reg_nopr} and Figure \ref{fig:success_vs_distance}, to compare against \crossmodalterm{} registration approaches~\cite{ren2022corri2p,zhou2024differentiable} we follow their setup by training on KITTI sequences 00-08 and evaluating on sequences KITTI-09 and 10.  Evaluation sequence statistics are given in Table \ref{tab:eval_info}.  We use the ground truth camera trajectory when constructing submaps for training and for evaluation on the KITTI360 sequences.  However, for evaluation on the KITTI dataset we construct submaps using the camera trajectory predicted by the visual odometry component of SLAM system OKVIS2~\cite{leutenegger2022okvis2} to demonstrate that \methodname{} can be straightforwardly implemented with existing out-of-the-box visual odometry solutions in the literature.

\subsection{Metrics and Evaluation}
\label{sec:metrics_evaluation}
For place recognition, following the established experimental setup in~\cite{komorowski2021egonn, vidanapathirana2022logg3d} we report the Recall@N (R@N) for N=$1,5$ for thresholds $5$m and $20$m.  For 6-DoF registration, previous works~\cite{komorowski2021egonn, zhou2024differentiable,xu2024c2l} report the mean relative rotation (RRE) and translation (RTE) error for \textit{successful} registrations only, where a successful registration is one where RRE $ \leq 5.0\degree$ and RTE $\leq 2.0 \text{m}$.  In addition, when evaluating unified place recognition and registration these works only report the registration accuracy (\ie{} percent of successful registrations) for top-1 positives retrieved by the place recognition module of the proposed approach.  This strategy results in inconsistent comparisons and inflated performance when benchmarking, since as noted by \cite{spectralgv,zhou2024differentiable} the low performances of failed registrations are removed from the reported result and the identity of the query-candidate pairings being registered is inconsistent between compared results.  Consequently, for fairer comparison we also report the registration accuracy and mean RRE and RTE for every possible positive candidate within a $20$m radius for each query submap, including values from unsuccessful registrations in our the reported RRE and RTE.  We refer to these two evaluation setups as the \textit{Top-1} and \textit{Comprehensive} setup respectively.\looseness=-1 %

\begin{table}[t]
    \centering 
    \begin{NiceTabular}{X||wc{0.41cm}wc{0.65cm}wc{0.65cm}|wc{0.41cm}wc{0.65cm}wc{0.65cm}}
        \CodeBefore
            \rowcolor{green!10}{7}
        \Body 
        \hline 
        \Block{2-1}{Method} & \Block{1-3}{KITTI-09*} & & &  \Block{1-3}{KITTI-10*} & & \\
        & Acc.\%  & RRE (\degree)  & RTE (m)   & Acc.\% & RRE (\degree) & RTE (m)  \\
        \hline 
        CorrI2P \cite{ren2022corri2p} & 8.1 & 13.59 & 11.11 & 9.0 & 9.44 & 10.15 \\
        VP2P \cite{zhou2024differentiable} & 6.9 & \underline{11.35} & 10.43 & 7.6 & \underline{8.16} & 10.00 \\
        EgoNN$^{\dag}$ \cite{komorowski2021egonn} &  50.7 & 22.30 & 7.35  & 57.1 & 27.86 & 6.59\\ %
        \hdashline 
        SOLVR-RANSAC & \underline{81.1} & 11.77 & \underline{3.46} & \underline{80.1} & {13.83} & \underline{3.46}  \\
        SOLVR (Ours)& \textbf{98.0} & \textbf{2.12} & \textbf{0.74} & \textbf{95.4} & \textbf{3.60} &  \textbf{0.98} \\
        \hline 
    \end{NiceTabular}
    
    \caption{6-DoF \crossmodalterm{} registration results for the \textit{Comprehensive} evaluation setup on KITTI-09 and 10}
    \label{tab:reg_nopr}
    \vspace{-5mm}
\end{table}

\subsection{Implementation Details} %
We train all of our experiments on a single NVIDIA Quadro RTX 6000 for 80 epochs, dropping the learning rate from $1e^{-3}$ to $1e^{-4}$ at 40 epochs, and set hyperparameters $m$, $d_{\text{thr}}$, $\tau$, $\alpha_{\text{pos}}^{\text{pr}}$ and $\alpha_{\text{pos}}^{\text{reg}}$ to $0.2$, $0.5$, $0.05$, $0.3$ and $0.1$ respectively.  
As baselines for unified re-localisation, we compare against EgoNN~\cite{komorowski2021egonn} trained on LiDAR-Visual data and \methodname{} with the registration proposed in \ref{sec:method_reg} replaced with RANSAC \cite{fischler1981random}.
For place recognition we also compare against the common non-learning based 3D baseline ScanContext~\cite{kim2018scan} and recently proposed LIPLoc~\cite{shubodh2024lip} and C2L-PR~\cite{xu2024c2l}.  Results from C2L-PR are reported only on KITTI-00, as the authors have not yet released their generated submaps for KITTI-360.  For \crossmodalterm{} registration, in addition to our baselines we also compare against recent state-of-the-art approaches CorrI2P~\cite{ren2022corri2p} and VP2P~\cite{zhou2024differentiable}, following their example in reporting results on sequences KITTI-09 and 10.

\section{Results}
\label{sec:results}

\subsection{Place Recognition Performance}

Table \ref{tab:pr_results} presents \methodname{} compared to our baselines and existing state-of-the-art methods in the literature.  \methodname{} out-performs competing \crossmodalterm{} approaches across the board, beating the nearest competitor by $16.6\%$, $18.1\%$ and $22.3\%$ in Recall@1 performance at the $5$m threshold on sequences KITTI-00, KITTI360-00 and KITTI360-09 respectively.  Looking at the specific competing approaches, we note that ScanContext \cite{kim2018scan} performs poorly due to significant differences in the bird's-eye-view encoding of our 3D submaps and corresponding LiDAR scans due to the limited field-of-view of the camera.  We also note that LipLoc \cite{shubodh2024lip} performs moderately well on the KITTI-00 and KITTI360-09 sequences but underperforms on sequence KITTI360-00.  We attribute this to the fact that our evaluation split on KITTI360-00 is composed almost entirely of reverse revisits, which is fundamentally challenging to an approach which only encodes the forward field-of-view of the vehicle.

\subsection{Registration Performance}
\label{sec:reg_performance}

\begin{figure}[t]
\begin{subfigure}[b]{0.49\columnwidth}
         \centering
         \includegraphics[width=\textwidth]{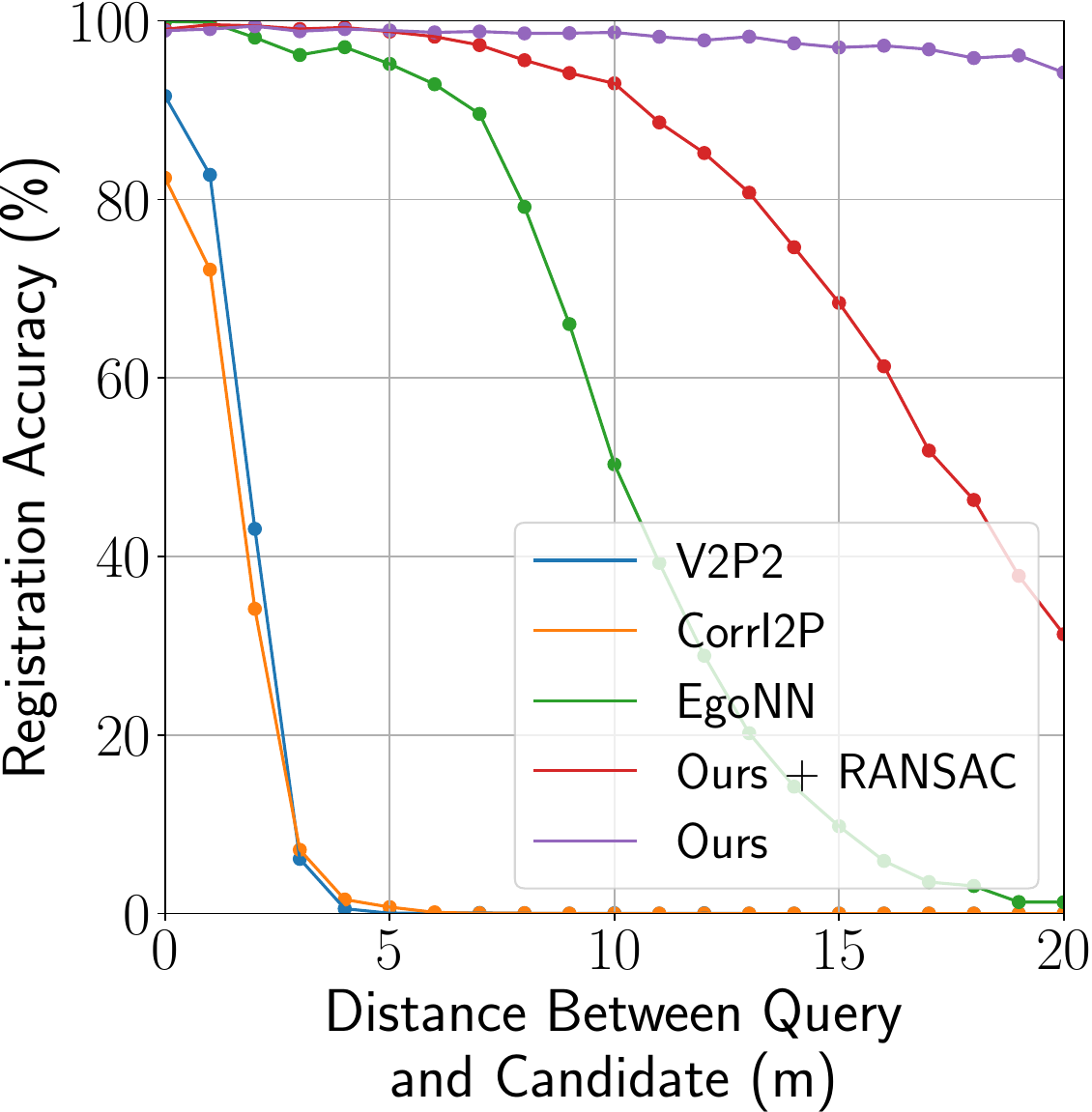}
         \caption{KITTI-09}
     \end{subfigure}
     \hfill 
    \begin{subfigure}[b]{0.49\columnwidth}
         \centering
         \includegraphics[width=\textwidth]{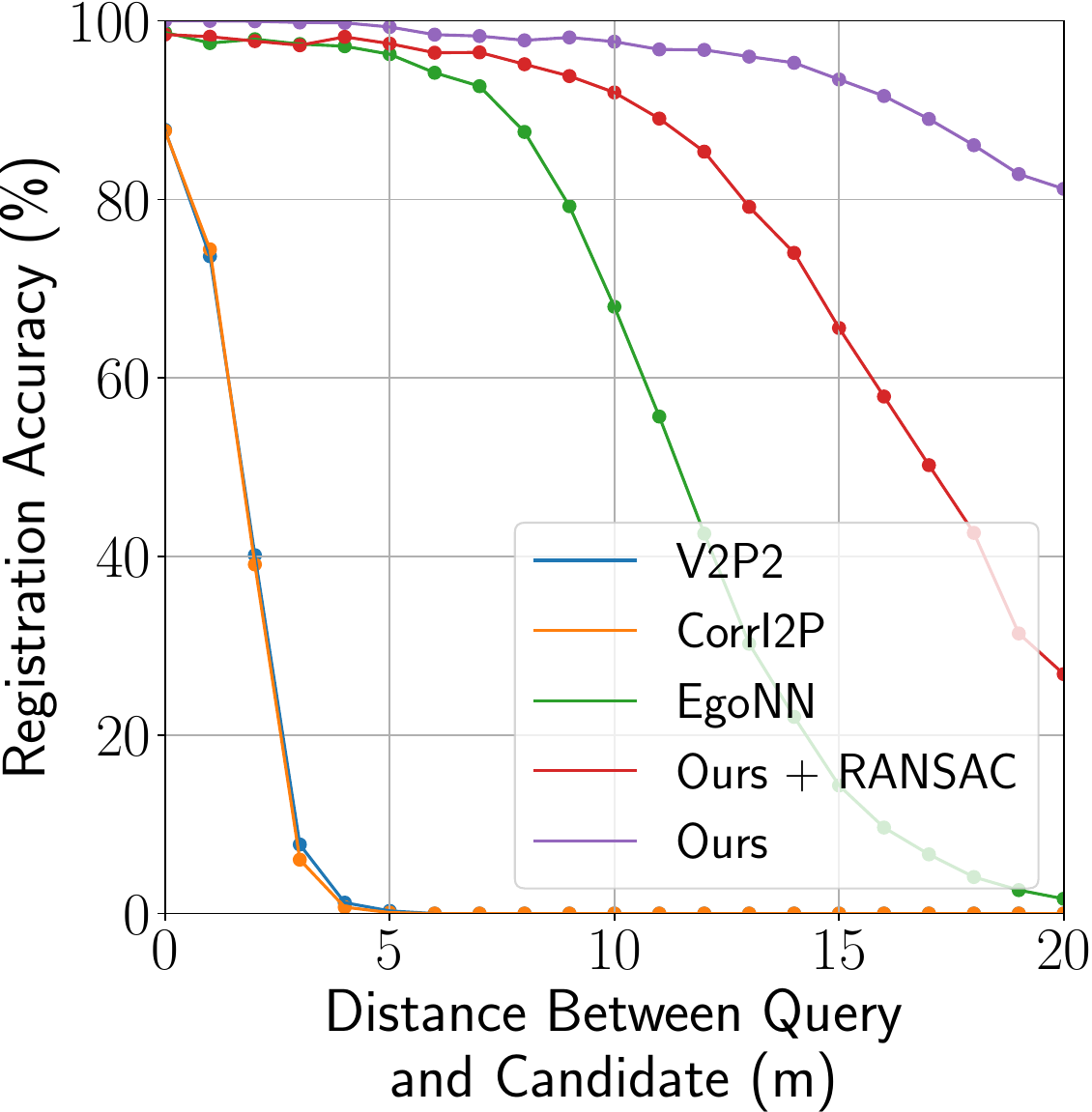}
         \caption{KITTI-10}
     \end{subfigure}

    \caption{6-DoF \crossmodalterm{} Registration accuracy versus distance between query and candidate pairs on sequences KITTI-09 and 10.  }
    
    \label{fig:success_vs_distance}
\end{figure}
\begin{figure}[t]
    \centering
    \includegraphics[width=0.95\linewidth]{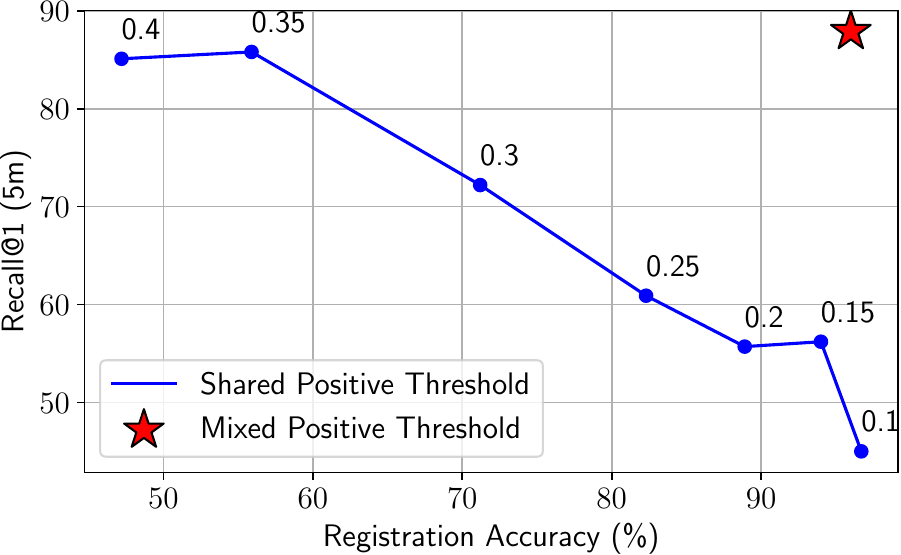}
    \caption{Recall@1 at 5m threshold vs. 6-DoF Registration accuracy for \crossmodalterm{} place recognition and registration for shared and mixed training thresholds on KITTI-00.  
    }
    \label{fig:recall_vs_success}
    \vspace{-5mm}
\end{figure}

Tables \ref{tab:reg_top1_results} and \ref{tab:reg_comp_results} report results for the \textit{Top-1} and \textit{Comprehensive} registration evaluation setups outlined in Section \ref{sec:metrics_evaluation}.  We note that while \methodname{} achieves state-of-the-art performance underneath both evaluation setups, the gap in performance between our approach and competing baselines widens dramatically moving from the \textit{Top-1} to \textit{Comprehensive} setups, with the difference in registration accuracy between \methodname{} and EgoNN \cite{komorowski2021egonn} increasing from 1.7\%, 16.9\% and 4.3\% to 29.4\%, 48.8\% and 39.7\% on sequences KITTI-00, KITTI360-00 and KITTI360-09 respectively.  We show in Figure \ref{fig:success_vs_distance} that a large part of this performance gap is attributable to \methodname{}'s greater robustness when registering difficult, far-away queries and candidates.  In particular we note that while state-of-the-art methods CorrI2P \cite{ren2022corri2p} and VP2P \cite{zhou2024differentiable} achieve fairly strong performance when the query image is situated directly in the center of the candidate LiDAR scan, their performance rapidly degrades as this assumption of centering weakens.  This is reflected in the results reported in Table \ref{tab:reg_nopr}, where both approaches report sub-$10\%$ registration accuracy on both sequences KITTI-09 and KITTI-10 under the \textit{Comprehensive} evaluation setup compared to \methodname{}'s accuracy of $98.0\%$ and $95.4\%$ on the two sequences respectively.  Finally, we note that in addition to consistently improving registration accuracy \methodname{} also achieves significantly faster runtime, with an average registration time of $2.59$ms compared to RANSAC at $39.50$ms with 10,000 iterations.

\begin{table}[t]
    \centering
    \begin{NiceTabularX}{\columnwidth}{X||cc|cc}
    \CodeBefore
            \rowcolor{green!10}{5}
        \Body 
    \hline 
        
         \Block{2-1}{Method} & \Block{1-2}{5m} && \Block{1-2}{20m} & \\
         & R@1 & R@5 & R@1 & R@5 \\
         \hline 
         Single Depth Image & 36.1 & 56.4 & 60.0 & 74.4   \\
         Naive Fusion & \underline{80.2} & \underline{89.1} & \underline{86.3} & \underline{93.6} \\
        SOLVR (Ours)& \textbf{87.9} & \textbf{94.0} & \textbf{96.4} & \textbf{96.8}  \\
    \hline 
    \end{NiceTabularX}
    \caption{\crossmodalterm{} place recognition performance for different submap generation strategies on KITTI-00.}
    \label{tab:ablation_mapgen_pr}
\end{table}

\begin{table}[t]
    \centering
    \begin{NiceTabularX}{\columnwidth}{X||ccc}
        \CodeBefore
            \rowcolor{green!10}{4}
        \Body 
        \hline 
         Method & Acc.\% & RRE ($\degree$) & RTE(m) \\
         \hline 
         Single Depth Image & 81.4 & 14.44 & 3.21  \\
         Naive Fusion & \underline{93.5} & \underline{5.61} & \underline{1.08}  \\
         SOLVR (Ours) & \textbf{96.0} & \textbf{3.39} & \textbf{1.02} \\
    \hline 
    \end{NiceTabularX}
    \caption{\crossmodalterm{} 6-DoF registration performance for different submap strategies on KITTI-00.}
    \label{tab:ablation_mapgen_reg}
    \vspace{-5mm}
\end{table}

\subsection{Ablation Studies}
In Tables \ref{tab:ablation_mapgen_pr} and \ref{tab:ablation_mapgen_reg} we compare the performance of our submap generation method with two different baselines: Single Depth Image, where we generate submaps with only the 3D projection of a single frame's depth image, and Naive Fusion, where we directly fuse the all of the 3D projections from the depth images in a local window without using the Bayesian update outlined in Section \ref{sec:submap_generation}.  We observe that \methodname{}'s submap generation significantly out-performs the Single Depth Image and Naive Concatenation baselines in place recognition performnce, with Recall@1 performance at the 5m threshold improved by $52\%$ and $7.7\%$ respectively, while also achieving moderate improvement in 6-DoF registration with a registration accuracy improvement of $14.6\%$ and $2.5\%$ over the respective baselines.

In Figure \ref{fig:recall_vs_success}, we plot the R@1 at 5m versus the registration accuracy on sequence KITTI-00 under the \textit{Comprehensive} evaluation setup for a range of positive training thresholds $\alpha_{\text{pos}}$, and compare it to the performance of a network trained with mixed positive thresholds $\alpha_{\text{pos}}^{\text{pr}}$ and $\alpha_{\text{pos}}^{\text{reg}}$.  We observe that there is a clear trade-off between place recognition and registration performance when a fixed positive threshold is used, whereas our approach of using a mixed threshold bypasses this trade-off and allows us to optimise performance on both tasks simultaneously, achieving both accuracte place recognition and robust 6-DoF registration from the same pipeline. %

\section{Conclusion}
\label{sec:conclusion}

In this work, we introduced \methodname{}, a pipeline for learning-based \crossmodalterm{} re-localisation. \methodname{} generates 3D submaps from camera images to align sensor modalities while retaining the benefits of LiDAR-based re-localisation without the added weight and cost of deploying a LiDAR sensor. By incorporating flexible positive thresholds during training and adapting recent advances in 3D registration, \methodname{} achieves state-of-the-art results in both place recognition and registration, offering faster and more reliable performance than existing methods. For future work, exploring multi-view stereo to reduce dependence on stereo cameras or leveraging cross-attention mechanisms for more context-aware registration are promising avenues for further research.

\section*{Acknowledgements}
Special thanks to Simon Boche, Jaehyung Jung and Sotiris Papatheodoru for their assistance and insight over the course of this work's development.  This work was supported by the Technical University of Munich, Data61 CSIRO, and the EU Horizon project DigiForest.

\balance{}

\bibliographystyle{./IEEEtran}
\bibliography{ref}

\begin{thebibliography}{10}
\providecommand{\url}[1]{#1}
\csname url@samestyle\endcsname
\providecommand{\newblock}{\relax}
\providecommand{\bibinfo}[2]{#2}
\providecommand{\BIBentrySTDinterwordspacing}{\spaceskip=0pt\relax}
\providecommand{\BIBentryALTinterwordstretchfactor}{4}
\providecommand{\BIBentryALTinterwordspacing}{\spaceskip=\fontdimen2\font plus
\BIBentryALTinterwordstretchfactor\fontdimen3\font minus \fontdimen4\font\relax}
\providecommand{\BIBforeignlanguage}[2]{{%
\expandafter\ifx\csname l@#1\endcsname\relax
\typeout{** WARNING: IEEEtran.bst: No hyphenation pattern has been}%
\typeout{** loaded for the language `#1'. Using the pattern for}%
\typeout{** the default language instead.}%
\else
\language=\csname l@#1\endcsname
\fi
#2}}
\providecommand{\BIBdecl}{\relax}
\BIBdecl

\bibitem{galvez2012bags}
D.~G{\'a}lvez-L{\'o}pez and J.~D. Tardos, ``Bags of binary words for fast place recognition in image sequences,'' \emph{IEEE Transactions on robotics}, vol.~28, no.~5, pp. 1188--1197, 2012.

\bibitem{hausler2021patch}
S.~Hausler, S.~Garg \emph{et~al.}, ``{Patch-NetVLAD}: Multi-scale fusion of locally-global descriptors for place recognition,'' in \emph{Proceedings of the IEEE/CVF conference on computer vision and pattern recognition}, 2021, pp. 14\,141--14\,152.

\bibitem{sarlin2019coarse}
P.-E. Sarlin, C.~Cadena \emph{et~al.}, ``From coarse to fine: Robust hierarchical localization at large scale,'' in \emph{Proceedings of the IEEE/CVF conference on computer vision and pattern recognition}, 2019, pp. 12\,716--12\,725.

\bibitem{uy2018pointnetvlad}
M.~A. Uy and G.~H. Lee, ``Pointnetvlad: Deep point cloud based retrieval for large-scale place recognition,'' in \emph{Proceedings of the IEEE conference on computer vision and pattern recognition}, 2018, pp. 4470--4479.

\bibitem{vidanapathirana2022logg3d}
K.~Vidanapathirana, M.~Ramezani \emph{et~al.}, ``{LoGG3D-Net: Locally guided global descriptor learning for 3d place recognition},'' in \emph{2022 International Conference on Robotics and Automation (ICRA)}.\hskip 1em plus 0.5em minus 0.4em\relax IEEE, 2022, pp. 2215--2221.

\bibitem{komorowski2021egonn}
J.~Komorowski, M.~Wysoczanska \emph{et~al.}, ``Egonn: Egocentric neural network for point cloud based 6dof relocalization at the city scale,'' \emph{IEEE Robotics and Automation Letters}, vol.~7, no.~2, pp. 722--729, 2021.

\bibitem{Cai2024VOLocVP}
X.~Cai, Y.~Wang \emph{et~al.}, ``Voloc: Visual place recognition by querying compressed lidar map,'' \emph{2024 IEEE International Conference on Robotics and Automation (ICRA)}, pp. 10\,192--10\,199, 2024.

\bibitem{xu2024c2l}
H.~Xu, H.~Liu \emph{et~al.}, ``C2l-pr: Cross-modal camera-to-lidar place recognition via modality alignment and orientation voting,'' \emph{IEEE Transactions on Intelligent Vehicles}, 2024.

\bibitem{zhao2023attention}
Z.~Zhao, H.~Yu \emph{et~al.}, ``Attention-enhanced cross-modal localization between spherical images and point clouds,'' \emph{IEEE Sensors Journal}, 2023.

\bibitem{zheng2023i2p}
S.~Zheng, Y.~Li \emph{et~al.}, ``I2p-rec: Recognizing images on large-scale point cloud maps through bird's eye view projections,'' in \emph{2023 IEEE/RSJ International Conference on Intelligent Robots and Systems (IROS)}.\hskip 1em plus 0.5em minus 0.4em\relax IEEE, 2023, pp. 1395--1400.

\bibitem{shubodh2024lip}
S.~Shubodh, M.~Omama \emph{et~al.}, ``Lip-loc: Lidar image pretraining for cross-modal localization,'' in \emph{Proceedings of the IEEE/CVF Winter Conference on Applications of Computer Vision}, 2024, pp. 948--957.

\bibitem{yin2021i3dloc}
P.~Yin, L.~Xu \emph{et~al.}, ``i3dloc: Image-to-range cross-domain localization robust to inconsistent environmental conditions,'' in \emph{17th Robotics: Science and Systems, RSS 2021}, 2021.

\bibitem{lee20232}
A.~J. Lee, S.~Song \emph{et~al.}, ``Lidar-camera loop constraints for cross-modal place recognition,'' \emph{IEEE Robotics and Automation Letters}, vol.~8, no.~6, pp. 3589--3596, 2023.

\bibitem{cattaneo2019cmrnet}
D.~Cattaneo, M.~Vaghi \emph{et~al.}, ``Cmrnet: Camera to lidar-map registration,'' in \emph{2019 IEEE intelligent transportation systems conference (ITSC)}.\hskip 1em plus 0.5em minus 0.4em\relax IEEE, 2019, pp. 1283--1289.

\bibitem{li2021deepi2p}
J.~Li and G.~H. Lee, ``Deepi2p: Image-to-point cloud registration via deep classification,'' in \emph{Proceedings of the IEEE/CVF Conference on Computer Vision and Pattern Recognition}, 2021, pp. 15\,960--15\,969.

\bibitem{zhou2024differentiable}
J.~Zhou, B.~Ma \emph{et~al.}, ``Differentiable registration of images and lidar point clouds with voxelpoint-to-pixel matching,'' \emph{Advances in Neural Information Processing Systems}, vol.~36, 2024.

\bibitem{r3loc2023}
M.~Ramezani, E.~Griffiths \emph{et~al.}, ``{Deep Robust Multi-Robot Re-Localisation in Natural Environments},'' in \emph{2023 IEEE/RSJ International Conference on Intelligent Robots and Systems (IROS)}, 2023, pp. 3322--3328.

\bibitem{bai2021pointdsc}
X.~Bai, Z.~Luo \emph{et~al.}, ``Pointdsc: Robust point cloud registration using deep spatial consistency,'' in \emph{Proceedings of the IEEE/CVF Conference on Computer Vision and Pattern Recognition}, 2021, pp. 15\,859--15\,869.

\bibitem{geiger2013kittidataset}
A.~Geiger, P.~Lenz \emph{et~al.}, ``{Vision meets robotics: The kitti dataset},'' \emph{The International Journal of Robotics Research}, vol.~32, no.~11, pp. 1231--1237, 2013.

\bibitem{liao2022kitti360}
Y.~Liao, J.~Xie \emph{et~al.}, ``{{KITTI}-360: A novel dataset and benchmarks for urban scene understanding in 2d and 3d},'' \emph{IEEE Transactions on Pattern Analysis and Machine Intelligence}, 2022.

\bibitem{ren2022corri2p}
S.~Ren, Y.~Zeng \emph{et~al.}, ``Corri2p: Deep image-to-point cloud registration via dense correspondence,'' \emph{IEEE Transactions on Circuits and Systems for Video Technology}, vol.~33, no.~3, pp. 1198--1208, 2022.

\bibitem{lepetit2009ep}
V.~Lepetit, F.~Moreno-Noguer \emph{et~al.}, ``Ep n p: An accurate o (n) solution to the p n p problem,'' \emph{International journal of computer vision}, vol.~81, pp. 155--166, 2009.

\bibitem{fischler1981random}
M.~A. Fischler and R.~C. Bolles, ``Random sample consensus: a paradigm for model fitting with applications to image analysis and automated cartography,'' \emph{Communications of the ACM}, vol.~24, no.~6, pp. 381--395, 1981.

\bibitem{sattler2016efficient}
T.~Sattler, B.~Leibe \emph{et~al.}, ``Efficient \& effective prioritized matching for large-scale image-based localization,'' \emph{IEEE transactions on pattern analysis and machine intelligence}, vol.~39, no.~9, pp. 1744--1756, 2016.

\bibitem{Sarlin2018FromCT}
P.-E. Sarlin, C.~Cadena \emph{et~al.}, ``From coarse to fine: Robust hierarchical localization at large scale,'' \emph{2019 IEEE/CVF Conference on Computer Vision and Pattern Recognition (CVPR)}, pp. 12\,708--12\,717, 2018.

\bibitem{du2020dh3d}
J.~Du, R.~Wang \emph{et~al.}, ``Dh3d: Deep hierarchical 3d descriptors for robust large-scale 6dof relocalization,'' in \emph{Computer Vision--ECCV 2020: 16th European Conference, Glasgow, UK, August 23--28, 2020, Proceedings, Part IV 16}.\hskip 1em plus 0.5em minus 0.4em\relax Springer, 2020, pp. 744--762.

\bibitem{zhang2023instaloc}
L.~Zhang, S.~Tejaswi~Digumarti \emph{et~al.}, ``Instaloc: One-shot global lidar localisation in indoor environments through instance learning,'' in \emph{Robotics: Science and Systems (RSS)}, 2023.

\bibitem{spectralgv}
K.~Vidanapathirana, P.~Moghadam \emph{et~al.}, ``{Spectral Geometric Verification: Re-Ranking Point Cloud Retrieval for Metric Localization},'' \emph{IEEE Robotics and Automation Letters}, vol.~8, no.~5, pp. 2494--2501, 2023.

\bibitem{zhang2024visual}
L.~Zhang, Y.~Tao \emph{et~al.}, ``Visual localization in 3d maps: Comparing point cloud, mesh, and nerf representations,'' \emph{arXiv preprint arXiv:2408.11966}, 2024.

\bibitem{xu2023unifying}
H.~Xu, J.~Zhang \emph{et~al.}, ``Unifying flow, stereo and depth estimation,'' \emph{IEEE Transactions on Pattern Analysis and Machine Intelligence}, 2023.

\bibitem{funk2021multi}
N.~Funk, J.~Tarrio \emph{et~al.}, ``Multi-resolution 3d mapping with explicit free space representation for fast and accurate mobile robot motion planning,'' \emph{IEEE Robotics and Automation Letters}, vol.~6, no.~2, pp. 3553--3560, 2021.

\bibitem{li2019usip}
J.~Li and G.~H. Lee, ``Usip: Unsupervised stable interest point detection from 3d point clouds,'' in \emph{Proceedings of the IEEE/CVF international conference on computer vision}, 2019, pp. 361--370.

\bibitem{komorowski2021minkloc3d}
J.~Komorowski, ``Minkloc3d: Point cloud based large-scale place recognition,'' in \emph{Proceedings of the IEEE/CVF Winter Conference on Applications of Computer Vision}, 2021, pp. 1790--1799.

\bibitem{komorowski2022improving}
------, ``Improving point cloud based place recognition with ranking-based loss and large batch training,'' in \emph{2022 26th international conference on pattern recognition (ICPR)}.\hskip 1em plus 0.5em minus 0.4em\relax IEEE, 2022, pp. 3699--3705.

\bibitem{knights2023geoadapt}
J.~Knights, S.~Hausler \emph{et~al.}, ``{GeoAdapt}: Self-supervised test-time adaptation in lidar place recognition using geometric priors,'' \emph{IEEE Robotics and Automation Letters}, vol.~9, no.~1, pp. 915--922, 2024.

\bibitem{spectralmatch}
M.~Leordeanu and M.~Hebert, ``{A spectral technique for correspondence problems using pairwise constraints},'' in \emph{Tenth IEEE International Conference on Computer Vision (ICCV'05) Volume 1}, 2005.

\bibitem{kim2018scan}
G.~Kim and A.~Kim, ``{Scan context: Egocentric spatial descriptor for place recognition within 3D point cloud map},'' in \emph{2018 IEEE/RSJ International Conference on Intelligent Robots and Systems}, 2018.

\bibitem{leutenegger2022okvis2}
S.~Leutenegger, ``Okvis2: Realtime scalable visual-inertial slam with loop closure,'' \emph{arXiv preprint arXiv:2202.09199}, 2022.

\end{thebibliography}

\end{document}